# Reducing Reasoning Costs - The Path of Optimization for Chain of Thought via Sparse Attention Mechanism

Libo Wang

Nicolaus Copernicus University / UCSI University

326360@o365.stud.umk.pl / 1002265630@ucsi.university.edu.my

*Abstract—* In order to address the chain of thought in the large language model inference cost surge, this research proposes to use a sparse attention mechanism that only focuses on a few relevant tokens. The researcher constructed a new attention mechanism and used GiantRabbit trained with custom GPTs as an experimental tool. The experiment tested and compared the reasoning time, correctness score and chain of thought length of this model and o1 Preview in solving the linear algebra test questions of MIT OpenCourseWare. The results show that GiantRabbit's reasoning time and chain of thought length are significantly lower than o1 Preview. It verifies the feasibility of sparse attention mechanism for optimizing chain of thought reasoning. Detailed architectural details and experimental process have been uploaded to Github, the link is:https://github.com/brucewang123456789/GeniusTrail.git.

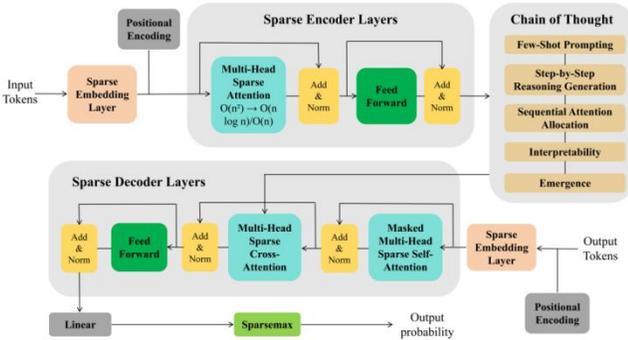

Fig. 1: An Innovative Transformer Architecture that Integrates Sparse Attention Mechanisms with Chain of Thought

## I. INTRODUCTION

Autoregressive modeling introduces chain of thought (CoT), adding a strategy to generate a series of intermediate reasoning steps before outputting content based on prompts (Mitra et al., 2024). The release of o1 Preview in 2024 has proven that resource optimization can maintain a high level of reasoning accuracy (Zhong et al., 2024). As shown in Figure 2, CoT most notably enhances the inference capabilities of large language models, which can be achieved by prompt engineering (Wei et al., 2022; Li et al., 2024).

The key to addressing the surge in reasoning cost caused by CoT lies in the computational complexity of the prune self-attention mechanism when dealing with long sequences (Jin et al., 2024). Since each token needs to be computationally related to all other tokens in the sequence, $O(n^2)$ is used for computational complexity (Condevaux & Harispe, 2023). As the sequence length increases, the computation will grow at the rate of quadrature. will grow at the rate of quadratic of n, which generates more tokens (Xiong et al., 2021; Zheng et al., 2024).

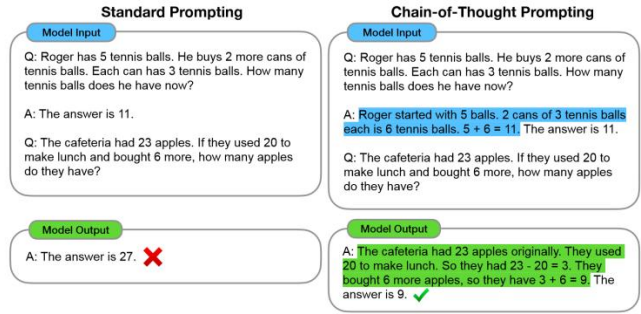

Figure 2: Chain of thought prompting enables large language models to tackle complex arithmetic, commonsense, and symbolic reasoning tasks (Adapted from Wei et al., 2022)

## II. SPARSE ATTENTION MECHANISM

Sparse attention is closely related to sparse coding in the human visual cortex, a principle that simulates the brain in the transformer to accurately extract information with low energy consumption and low time cost (Zheng et al., 2023). By simulating sparse coding in the visual cortex of the human brain, this mechanism can theoretically reduce computational complexity and maintain model performance in information processing (Olshausen & Field, 2004; Rego et al., 2023). In view of the optimization of information transfer and computational efficiency, it is supported by graph theory (Li et al., 2023). This mechanism aims to reduce the complexity of the attention computation by limiting each attention to only a small number of lexemes that are highly relevant (Frantar & Alistarh, 2023).

### A. Sparse Transformer Architecture

Firstly, tokens are inputted and initially embedded into sparse vectors by sparse embedding layer, which effectively reduces the computational burden of high-dimensional data. Next, residual linking and regularization are performed through the add & norm layer to ensure the stability of the model during the deep learning process (Vaswani, 2017). The data then flows through the feed forward that is set up as a two-layer fully connected network on a location-by-location basis (Geva et al., 2022). After the input has been processed at the coding level, the chain of thought module is responsible for generating intermediate inference steps to split the complex problem into smaller logical units (Khot et al., 2022).

In the sparse decoder layer, masked multi-head sparse self-attention is first used to ensure that the model can only see previous outputs. The data goes through the add & norm layer to maintain the gradient stability. Immediately after that, the data enters multi-head sparse cross attention to introduce CoT reasoning attention operation to the output of coding layer.

After the feed forward network and two residual connections, the decoder performs a linear transformation of the output through the linear layer. It is worth noting that the research chose sparsemax instead of softmax as the activation function to generate word probability distributions (Martins & Astudillo, 2016).

B. Proposed Algorithms

First, the embedding matrix $E \in R^{V \times D}$, where $V$ represents the size of the vocabulary and $D$ represents the embedding dimension. For the input token index $x \in R^{B \times N}$, where $B$ represents the batch size and $N$ represents the sequence length.

$$\text{Embed}(x) = E[x] \in \mathbb{R}^{B \times N \times D}$$

To apply sparsity in the embedding representation to reduce the computational complexity, this research introduces the sparsity mask $M \in \{0,1\}^D$. The generation is based on the sparsity factor recorded as $\alpha$ to control the embedding dimensions. Specifically, for the sparsity mask $M$, its elements $M_i$ are defined as the following mathematical representation, where the set $S$ represents the selected active dimensions, determined by the sparsity factor $\alpha$.

$$M_i = 1 \cdot (i \in S) + 0 \cdot (i \notin S)$$

Applying the sparsity mask $M$ to the embedding representation process can be expressed mathematically as follows. $\odot$ represents element-level multiplication operations, and broadcast operations in batch and sequence dimensions.

$$\text{SparseEmbed}(x) = \text{Embed}(x) \odot M$$

The core of the current algorithm is based on the principle of sparse attention, which limits each label to only focus on a part of the most relevant tokens to reduce the amount of attention required to be calculated for each token (Guan et al., 2022; Yun et al., 2024). Different from the traditional self-attention mechanism, it does not need to calculate the correlation of each token with all other tokens, reducing the computational complexity from $O(n)^2$ to $O(n)$ or $n*log(n)$ (Kitaev et al., 2020; Treviso et al., 2021).

$$\text{SparseAttention}(Q,K,V,S) = \text{sparsemax}(\frac{QK^T}{\sqrt{d_k}} + M)V$$

$Q$, $K$, $V$ represent query, key and value matrices respectively. $S$ represents sparse mode that defines each token should focus on; $M$ represents the mask matrix. With large negative values such as -oo at locations where attention is not allowed, it effectively zeroes out some attention weights. Sparsemax is an activation function that converts attention scores into sparse probability distributions. Less important tokens are actually assigned the weight of zero.

The researcher first consider the output of the sparse coding layer as the input representation of CoT, denoted as SparseEmbed($x$). Its reasoning process is mathematically expressed as the following series of step-by-step updates:

$$R_t = f_t(\text{SparseEmbed}(x), C_{t-1})$$

$R_t$ represents the reasoning state at reasoning step t; $f_t$ represents the transformation function of each step, which includes a sparse multi-head attention layer and a sparse feed-forward neural network; $C_{t-1}$ represents the context information generated by the previous step of reasoning.

The researcher apply a sparsity mask after each reasoning step to retain the most representative features. It sparses the reasoning state $R_t$:

$$R_t^{sparse} = R_t \odot M_t$$

$M_t$ is a sparsity mask dynamically generated given the importance of intermediate features at each step of inference. Specifically, assuming that the output of the sparse encoder is $H_e$, the cross-attention weight is calculated as follows:

$$A_{cross} = \text{sparsemax}(\frac{QK^T}{\sqrt{d_k}} + M_{cross})$$

$Q$ represents the query vector from the decoder; $K$ represents the key vector from the encoder; $M_{cross}$ represents the sparsity mask in cross attention to limit the attention computation to only focus on the most important inputs.

In the calculation of sparse self-attention, this part ensures that each token in the decoder can only pay attention to the previously generated tokens, while following the sparsity principle:

$$A_{self} = \text{sparsemax}(\frac{QK^T}{\sqrt{d_k}} + M_{self})$$

$M_{self}$ is represented as a causal mask to prevent future tokens in the decoder from receiving attention, and enforces sparsity to limit the attention scope of each token.

Appropriate masking strategy selection improves the computational efficiency and reasoning accuracy of the multi-head sparse attention layer. Dynamic masking strategy generates a sparse matrix $M$ based on the correlation of input tokens, filters relevant tokens and ignores irrelevant parts before $QK^T$ calculation.

The final decoding output can be obtained by weighting and summing the attention of these two parts:

$$H_d = \text{FFN}(A_{self} \cdot V + A_{cross} \cdot H_e) \odot M_d$$

FFN stands for feedforward neural network. $M_d$ is the sparsity mask applied to the final output to ensure that the sparsity features in the output stage are preserved.

Since the Add & Norm layer continues the components of the standard transformer, it does not require major algorithm modifications to adapt to sparsity, so this layer does not have to come up with a new mathematical representation of the algorithm.

III. EXPERIMENT

Based on figure 2, this research performed prompt engineering in GPTs to train the model "GiantRabbit" with the sparse attention mechanism as the core. Its encoder-decoder architecture is different from the running process of the original GPTs series. In order to verify the effectiveness of this mechanism in optimizing reasoning costs, the researcher compared the performance of GiantRabbit and o1 Preview on three indicators: reasoning time, correctness score and CoT length. The experiment uses 9 sample questions from Stanford University's MATH 113 Linear Algebra (Exam 1) in the fall of 2018 as test data through MIT OpenCourseWare. This public exam is authorized for non-commercial educational use. Because o1 Preview cannot directly upload documents, the researcher manually entered the test questions into GiantRabbit and o1 Preview respectively. In addition, the researchers took average values and excluded abnormal data caused by API or network delays.

## IV. RESULT

According to the results in Figure 3, o1 Preview and GiantRabbit show significant differences on nine questions.

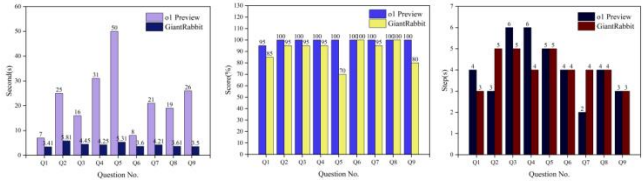

Fig. 3: Comparison between o1 preview and GiantRabbit in reasoning time, correctness score and CoT length

As evidenced by the data, o1 Preview's accuracy remained stable between 95% and 100%, but its reasoning time was significantly longer, ranging from 7 seconds to 50 seconds. In contrast, although GiantRabbit's accuracy fluctuates slightly (70%-100%), it shows a significant efficiency advantage in inference time, taking only 3.5 seconds to 5.8 seconds. In addition, GiantRabbit requires fewer reasoning steps on average, especially when dealing with questions with simpler reasoning steps (such as Q6 and Q9), demonstrating its advantage in simplifying the reasoning process. Figure 4 further shows the fluctuations and performance differences between o1 Preview and GiantRabbit during the problem solving process, clearly reflecting GiantRabbit's superior performance in reasoning time efficiency, while showing its relative trade-off in accuracy in a few situations.

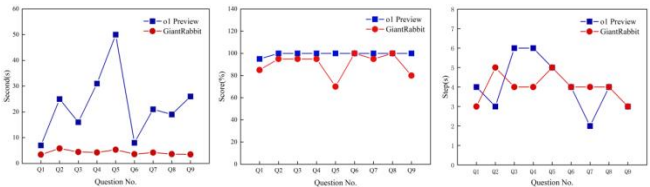

Figure 4: Comparison between o1 preview and GiantRabbit in performance stability

## V. RESULTS

Although GPTs that simulate Python is used as an experimental executor and has powerful data processing and analysis capabilities, it has inherent flaws in large language models, such as implicit biases in the reasoning process and limited accuracy in processing complex data. This requires researchers to conduct multiple experiments to verify the sample efficiency, adaptation speed, robustness, policy stability and domain gap, and retain the optimal data results. Table 1 shows the results of quantitative analysis for mechanical synthesis data.

Table 1. Mechanical metrics results

| Frame work | Sample Efficiency | Adaptation Speed | Robustness | Policy Stability | Domain Gap |
|---|---|---|---|---|---|
| ADR | 0.568 | 0.032 | 0.599 | 0.485 | 0.047 |
| TDR | 0.745 | 0.041 | 0.946 | 0.717 | 0.074 |

Table 2 is the results of quantitative analysis for environmental synthesis data.

Table 2. Environmental metrics results

| Frame work | Sample Efficiency | Adaptation Speed | Robustness | Policy Stability | Domain Gap |
|---|---|---|---|---|---|
| ADR | 0.479 | 0.018 | 0.543 | 0.531 | 0.043 |
| TDR | 0.607 | 0.026 | 0.821 | 0.673 | 0.063 |

As indicated by mechanical synthesis data, TDR's sample efficiency reaches 0.745, which is significantly superior to ADR's 0.568, which reflects its ability to achieve more efficient learning and adaptation with fewer samples; in terms of policy stability indicators, TDR The value is 0.717, which is better than ADR's 0.485. It shows that TDR maintains high consistency and reliability under dynamic conditions. In terms of robustness, TDR reaches 0.946, which is ahead of ADR's 0.599. However, its adaptation speed and domain gap index values are not significantly different from ADR. It situation also occurs in the analysis of environmental synthetic data, where TDR also shows better performance than ADR. The sample efficiency of TDR is 0.607, which is higher than ADR's 0.479. In terms of policy stability, TDR's 0.673 is also significantly better than ADR's 0.531. And the robustness of TDR is 0.821 which is still higher than ADR's 0.543. Similarly, in terms of adaptation speed and domain gap indicators, although TDR is higher than ADR, the advantage is not obvious.

## VI. LIMITATIONS

Currently, OpenAI updates the knowledge base until October 2023, GiantRabbit is trained on GPT-4Turbo. This causes certain interference to the experimental results because it is different from the o1 Preview configuration. In addition, owing to using commercial language models may cause reasoning time to be affected by API latency and network traffic.

## VII. CONCLUSION

This research's encoder-decoder transformer architecture based on the sparse attention mechanism shows excellent performance in chain of thought reasoning. The researcher compared the performance of GiantRabbit and o1 Preview on the MIT OpenCourseWare linear algebra test questions. The results showed that GiantRabbit has a significant advantage in inference time and number of inference steps, although there is a slight trade-off in accuracy. Experimental results prove that the sparse attention mechanism effectively reduces the cost of chain of thought reasoning, providing experience for training LLMs combined with sparse attention in the future.

APPENDIX 1

This research selects and excerpts exam questions from Exam 1 of MIT OpenCourseWare - Linear Algebra, and uses o1 Preview and GiantRabbit to answer them respectively. Since the test questions contain matrix operations, the researcher converted them into computer language that can be recognized by the model without changing the content of the test questions. It is available non-commercially for the experimental purposes of this research under the terms of the Creative Commons Attribution-NonCommercial-ShareAlike (CC BY-NC-SA) license. The copyright and originality of this test question also belongs to MIT OCW. The following are the original Exam 1 questions.

Section A: 18.06 Quiz   March 1, 2010   Professor Strang

1. Forward elimination changes Ax = b to a row reduced Rx = d:  the complete solution is

$$x = \begin{bmatrix} 4 \\ 0 \\ 0 \end{bmatrix} + c_1 \begin{bmatrix} 2 \\ 1 \\ 0 \end{bmatrix} + c_2 \begin{bmatrix} 5 \\ 0 \\ 1 \end{bmatrix}$$

(a) Wat is the 3 by 3 reduced row echelon matrix R and what is d?

(b) If the process of elimination subtracted 3 times row 1 from row 2 and then 5 times row 1 from row 3, what matrix connects R and d to the original A and b? Use this matrix to find A and b.

2. Suppose A is the matrix

$$A = \begin{bmatrix} 0 & 1 & 2 & 2 \\ 0 & 3 & 8 & 7 \\ 0 & 0 & 4 & 2 \end{bmatrix}.$$

(a) Find all special solutions to Ax = 0 and describe in words the whole nullspace of A.

(b) Describe the column space of this particular matrix A.  "All combinations of the four columns" is not a sufficient answer.

(c) What is the reduced row echelon form R* = rref(B) when B is the 6 by 8 block matrix using the same A?

$$B = \begin{bmatrix} A & A \\ A & A \end{bmatrix}$$

3. Circle the words that correctly complete the following sentence:

(a) Supose a 3 by 5 matrix A has rank r = 3. Then the equation Ax = b (always / sometimes but not always) has (a unique solution / many solutions / no solution).

(b) What is the column space of A? Describe the nullspace of A.

4. Suppose that A is the matrix

$$A = \begin{bmatrix} 2 & 1 \\ 6 & 5 \\ 2 & 4 \end{bmatrix}.$$

(a) Explain in words how knowing all solutions to Ax = b decides if a given vector b is in the column space of A.

(b) Is the vector b = in the column space of A?

Section B: Conversion of Computer Language

Each sub-question within an exam question can be reasoned about independently and is therefore numbered from 1 to 9 on the same dimension. In order to ensure the rigor of the experiment, no changes will be made to the content of the test questions, nor any additional special processing will be done to the execution. Since the test questions do not have images, Table 1 shows the conversion of questions with matrix operations into computer language that can be recognized by the model.

Table 1 - Adapted Questions of Exam 1

| No. | | Adapted Questions |
|---|---|---|
| 1 | Forward elimination changes Ax = b to a row reduced Rx = d: the complete solution is x = [4, 0, 0]$^T$ + c1 * [2, 1, 0]$^T$ + c2 * [5, 0, 1]$^T$ | What is the 3 by 3 reduced row echelon matrix R and what is d? |
| 2 | | If the process of elimination subtracted 3 times row 1 from row 2 and then 5 times row 1 from row 3, what matrix connects R and d to the original A and b? Use this matrix to fnd A and b. |
| 3 | Suppose A is the matrix A = [ [0, 1, 2, 2], [0, 3, 8, 7], [0, 0, 4, 2] ]. | Find all special solutions to Ax=0 and describe in words the whole nullspace of A. |
| 4 | | Describe the column space of this particular matrix A. "All combinations of the four columns" is not a sufficient answer |
| 5 | | What is the reduced row echelon form R* = rref(B) when B is the 6 by 8 block matrix B = [ A A ]  [ A A ] using the same A? |
| 6 | Rank and Solutions | Suppose a 3x5 matrix A has rank r=3. Then the equation Ax=b (circle the correct options) (always / sometimes but not always) has (a unique solution / many solutions / no solution). |
| 7 | | What is the column space of A? Describe the nullspace of A. |
| 8 | Suppose that A is the matrix A = [ [2, 1], [6, 5], [2, 4] ]. | Explain in words how knowing all solutions to Ax = b decides if a given vector b is in the column space of A. |
| 9 | | Is the vector b = [8, 28, 14] in the column space of A? |